\def\paperlanguage{}
\pdfoutput=1 

\documentclass[letterpaper, 10 pt, conference]{ieeeconf}  

\usepackage{bm}
\usepackage{cite}
\include{preamble}

\newcommand{\ctext}[1]{\raise0.2ex\hbox{\textcircled{\scriptsize{#1}}}}

\IEEEoverridecommandlockouts                              

\title{\LARGE \textbf
  {
    \switchlanguage%
    {%
      Design Optimization of Wire Arrangement with Variable Relay Points in Numerical Simulation for Tendon-driven Robots
    }%
    {%
      シミュレーション上での可変経由点を含む腱駆動ロボットの\\ワイヤ配置設計最適化
    }%
  }
}

\author{Kento Kawaharazuka$^{1}$, Shunnosuke Yoshimura$^{1}$, Temma Suzuki$^{1}$, Kei Okada$^{1}$, and Masayuki Inaba$^{1}$
  \thanks{$^{1}$ The authors are with the Department of Mechano-Informatics, Graduate School of Information Science and Technology, The University of Tokyo, 7-3-1 Hongo, Bunkyo-ku, Tokyo, 113-8656, Japan.
    {\texttt\small [kawaharazuka, yoshimura, t-suzuki, k-okada, inaba]@jsk.t.u-tokyo.ac.jp}
  }
}
\begin{document}

\maketitle
\thispagestyle{empty}
\pagestyle{empty}

\begin{abstract}
  \switchlanguage%
  {%
    One of the most important features of tendon-driven robots is the ease of wire arrangement and the degree of freedom it affords, enabling the construction of a body that satisfies the desired characteristics by modifying the wire arrangement.
    Various wire arrangement optimization methods have been proposed, but they have simplified the configuration by assuming that the moment arm of wires to joints are constant, or by disregarding wire arrangements that span multiple joints and include relay points.
    In this study, we formulate a more flexible wire arrangement optimization problem in which each wire is represented by a start point, multiple relay points, and an end point, and achieve the desired physical performance based on black-box optimization.
    We consider a multi-objective optimization which simultaneously takes into account both the feasible operational force space and velocity space, and discuss the optimization results obtained from various configurations.
  }%
  {%
    腱駆動型の身体は可変剛性や冗長駆動等の利点を多数有するが, その中でもワイヤ配置の容易さやその自由度は大きな特徴の一つである.
    ワイヤ配置を工夫することで, 所望の特性を満たす身体が構成可能であり, 様々なワイヤ配置最適化手法が提案されてきた.
    しかし, それらはワイヤの関節に対するモーメントアームを一定と仮定していたり, 複数の関節に渡り, かつ折返しや多様な経由を含むワイヤ配置を無視していたりと, その問題を単純化してきた.
    そこで本研究では, ワイヤを始点, 複数の中継点, 終止点で表現する, より自由度の高いワイヤ配置の最適化問題を定式化し, ブラックボックス最適化に基づき所望の身体性能を実現する.
    発揮可能手先力空間と発揮可能手先速度空間を同時に考慮する多目的最適化について考え, 多様な設定から得られた最適化結果を考察する.
  }%
\end{abstract}

\section{INTRODUCTION}\label{sec:introduction}
\switchlanguage%
{%
  A variety of tendon-driven robots have been constructed so far \cite{endo2019superdragon, yoshimura2023kangaroo, kawaharazuka2019musashi}.
  These have various advantages such as variable stiffness control \cite{kobayashi1998tendon} and robust response to wire breakage \cite{kawaharazuka2022redundancy}.
  Among the advantages, we focus on the ease of wire arrangement and the degree of freedom it affords in this study.
  Compared to axis-driven robots, tendon-driven robots can freely select the start, relay, and end points of wires, and can easily construct bodies with various configurations \cite{kawaharazuka2019musashi}.
  By modifying the wire arrangement, it is possible to construct a body that satisfies the desired characteristics, and various wire arrangement design optimization methods have been proposed so far.
  \cite{pollard2002arrangement} numerically optimizes muscle Jacobian and pulley radius for robot fingers to ensure a torque space equivalent to that of humans.
  \cite{rayne2018arrangement} exploratively optimizes the distance between joints and wires to enlarge the feasible joint angle space for a continuum robot.
  \cite{dong2018arrangement} optimizes the spacing of fingers, wire pulley radius, and pulley spacing based on genetic algorithm for a two-fingered hand.
  \cite{roozing2016arrangement} numerically optimizes the wire pulley radius, elasticity, and pretension for the design of asymmetric compliance actuators.
  \cite{hamida2021arrangement} optimizes the wire attachment position for Cable-Driven Parallel Robots (CDPR) based on evolutionary algorithm.
  \cite{jamshidifar2017arrangement} numerically optimizes the tension and wire attachment position for CDPR to optimize its stiffness.
  \cite{zhong2022arrangement} numerically optimizes the start and end point positions of muscles for a musculoskeletal robot with the attached link of each point fixed.
  \cite{asaoka2012arrangement} optimizes the presence or absence of moment arm of each muscle for each joint for a musculoskeletal robot with brute force search.
  \cite{kawaharazuka2021redundancy} optimizes muscle Jacobian for a musculoskeletal robot by genetic algorithm.
}%
{%
  これまで多様な腱駆動型のロボットが構築されてきた\cite{endo2019superdragon, gravato2010ecce1, kawaharazuka2019musashi}.
  これらは冗長なワイヤ配置と非線形弾性要素に基づく可変剛性制御\cite{kobayashi1998tendon}や, 冗長性に基づくワイヤ破断へのロバストな対応\cite{kawaharazuka2022redundancy}など, 多様な利点を有する.
  その中でも, 本研究ではワイヤ配置の容易さとその自由度に着目する.
  腱駆動型のロボットは, 軸駆動型のロボットに比べワイヤの始点・中継点・終止点を自由に選ぶことができ, 多様な身体を容易に構成できる\cite{kawaharazuka2019musashi}.
  ワイヤ配置を工夫することで, 所望の特性を満たす身体が構成可能であり, これまで様々なワイヤ配置設計最適化手法が提案されてきた.
  \cite{pollard2002arrangement}はロボットフィンガについて, 人間と同等なトルク空間を確保するために, モーメントアーム一定の条件下で筋長ヤコビアンとプーリの半径を数値的に最適化している.
  \cite{rayne2018arrangement}はcontinuumロボットについて, feasible joint angleの空間をenlargeするために, 関節とワイヤの距離を探索的に最適化している.
  \cite{dong2018arrangement}は二指ハンドの間隔やプーリサイズ, プーリ間隔を含む腱配置の最適化を遺伝的アルゴリズムに基づき行っている.
  \cite{ansari2018optimization}はソフトロボットアームにおけるワイヤの張力や流体アクチュエータの圧力を強化学習により設計している.
  \cite{roozing2016arrangement}は非対称コンプライアンスアクチュエータの設計に向けてワイヤのプーリ半径や弾性, プリテンションを数値的に最適化している.
  \cite{hamida2021arrangement}はCDPRについて, そのワイヤ取り付け位置を進化計算に基づき最適化している
  \cite{jamshidifar2017arrangement}はCDPRについて, その剛性最適化に向けた張力やワイヤ取り付け位置を数値的に最適化している.
  \cite{zhong2022arrangement}は筋骨格型のロボットについて, 各経由点の取り付けリンクを固定したうえで, 筋の起始点と終止点の位置を数値的に最適化している.
  \cite{asaoka2012arrangement}は筋骨格型のロボットについて, モーメントアーム固定の条件下で各筋の各関節に対するモーメントアームの有無を全探索で最適化している.
  \cite{kawaharazuka2021redundancy}は筋骨格型のロボットについて, どの筋が切れても動作継続可能な筋長ヤコビアンをモーメントアーム一定の条件下で遺伝的アルゴリズムにより最適化している.
}%

\begin{figure}[t]
  \centering
  \includegraphics[width=0.95\columnwidth]{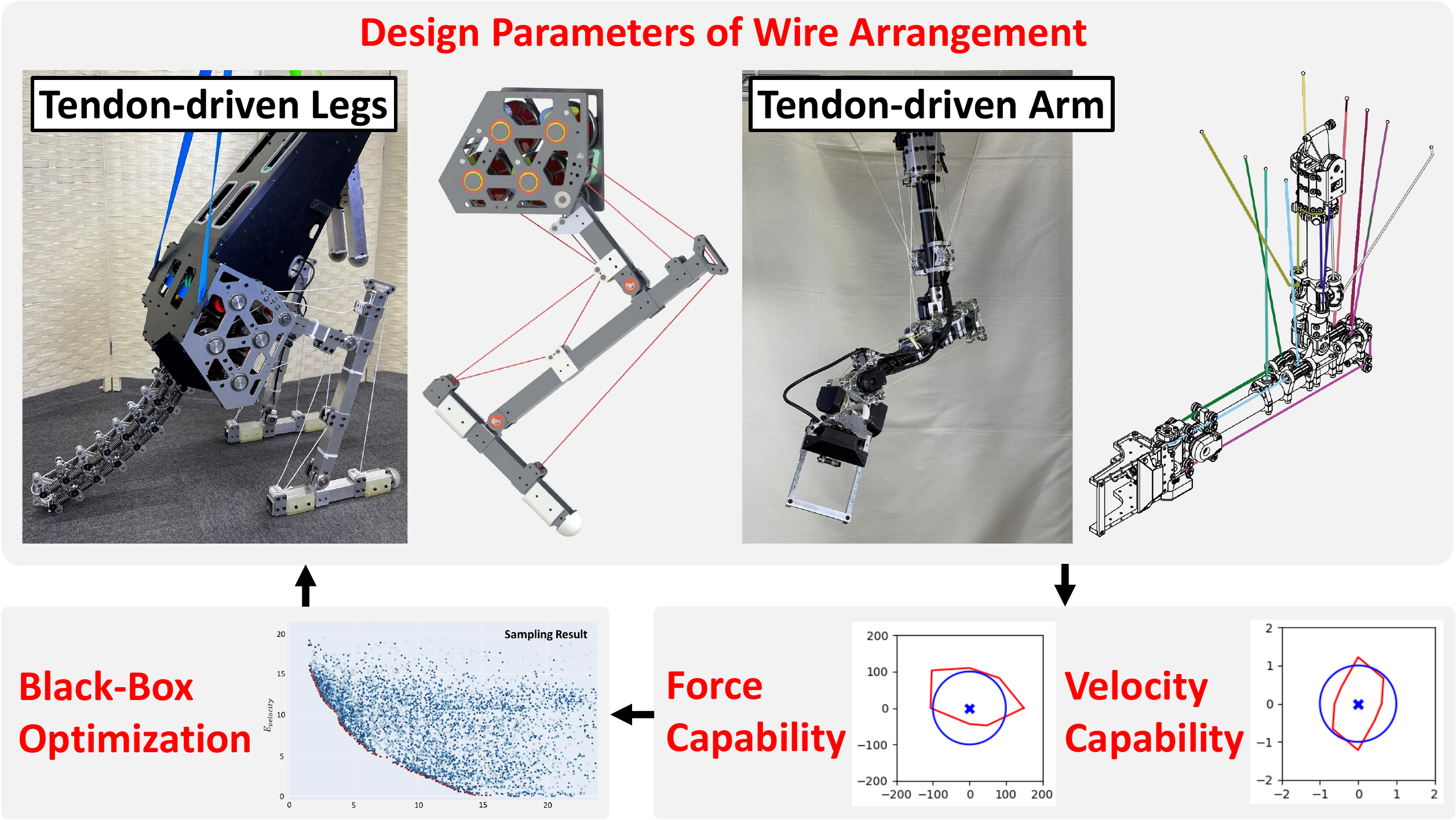}
  \vspace{-1.0ex}
  \caption{The concept of this study. We prepare design parameters of wire arrangement, calculate the evaluation value for target and feasible force / velocity regions regarding each design, and obtain the design parameter with the best performance by multi-objective black-box optimization.}
  \label{figure:concept}
  \vspace{-3.0ex}
\end{figure}

\switchlanguage%
{%
  Here, the wire arrangement configuration is mainly divided into two types: a type with constant moment arm using pulleys \cite{pollard2002arrangement, dong2018arrangement, roozing2016arrangement, asaoka2012arrangement, kawaharazuka2021redundancy}, and a type in which the moment arm changes depending on the joint angle by expressing the wire route with its start, relay, and end points \cite{rayne2018arrangement, hamida2021arrangement, jamshidifar2017arrangement, zhong2022arrangement}.
  Note that for the type with constant moment arm, the moment arm can be designed directly by the pulley radius, and it is easy to guarantee the moment arm at various joint angles, but the ease of wire arrangement is lost because it is difficult to obtain large moment arm due to the large pulleys placed at all joints.
  In terms of optimization methods, there are two types: those that deal mainly with continuous values and can be optimized analytically \cite{pollard2002arrangement, roozing2016arrangement, jamshidifar2017arrangement, zhong2022arrangement}, and those that use black-box optimization for discrete values or other complex settings \cite{rayne2018arrangement, dong2018arrangement, hamida2021arrangement, asaoka2012arrangement, kawaharazuka2021redundancy}.

  On the other hand, these studies have either focused on optimizing constant moment arms or only the start and end points of wires, disregarding complex wire configurations that span multiple joints and include bends with various relay points.
  Also, research on complex optimization involving both continuous and discrete values from multiple objective functions is scarce.
  If we can explore the presence or absence of wire bends, links to which the relay points are attached, and changes in the attached positions, it should be possible to create body configurations that can appropriately perform a wider variety of tasks.
  Therefore, we propose a wire arrangement design optimization method that takes into account variable relay points.
  We perform multi-objective black-box optimization by setting up a problem in which each wire can freely choose which position of each link it passes through, and by setting the realization of target operational force and velocity spaces as objective functions (\figref{figure:concept}).
  We show that various target operational force and velocity spaces can be realized by appropriately selecting the wire relay points, and discuss how the performance varies with the number of wires and relay points.
  We also show that the performance of this study is equivalent to or better than that of a configuration with constant moment arm using pulleys, while maintaining the degree of freedom of wire arrangement and large moment arm.

}%
{%
  その適用先であるワイヤ配置構成は主に, プーリを使用したモーメントアームを一定とするタイプ\cite{pollard2002arrangement, dong2018arrangement, roozing2016arrangement, asaoka2012arrangement, kawaharazuka2021redundancy}, ワイヤをその始点・中継点・終止点により表現しモーメントアームが関節角度に依存して変化するタイプに分けられる\cite{rayne2018arrangement, hamida2021arrangement, jamshidifar2017arrangement, zhong2022arrangement}.
  なお, モーメントアーム一定のタイプは, プーリ径により直接モーメントアームを設計でき, 多様な関節角度でモーメントアームを確保しやすい反面, 全関節に大きなプーリを配置するためワイヤ配置の容易さが失われ, かつモーメントアームも大きく取りにくいという欠点がある.
  また最適化手法について, 主に連続値のみを扱い解析的に最適化が可能な問題設定\cite{pollard2002arrangement, roozing2016arrangement, jamshidifar2017arrangement, zhong2022arrangement}と, 離散値やその他複雑な設定からブラックボックス最適化を用いる問題設定に分けられる\cite{rayne2018arrangement, dong2018arrangement, hamida2021arrangement, asaoka2012arrangement, kawaharazuka2021redundancy}.
  一方で, これらの研究は複数の関節に跨り, ワイヤの折返しや多様な経由点を含むような複雑なワイヤ配置を無視している.
  ワイヤの折返しの有無や, 経由点を取り付けるリンクや取り付け位置の変化を探索することができれば, より多様なタスクを適切にこなす身体構成が可能なはずである.
  そこで我々は, 可変ワイヤ経由点を考慮したワイヤ配置設計最適化手法を提案する.
  各ワイヤがどのリンクのどの位置を経由するかを自由に選択可能な問題設定を行い, これを目標手先力空間への合致と目標手先速度空間への合致を目的関数として, 多目的ブラックボックス最適化を行う.
  本手法によりワイヤ経由点を巧みに選ぶことで, 多様な目標手先力空間・目標手先速度空間を実現可能であることを示し, ワイヤ数や経由点数によるその性能の変化について議論する.
  また, ワイヤ配置の自由度や大きなモーメントアームを保ちながら, プーリ型でモーメントアーム一定なワイヤ駆動機構と同等かそれ以上の性能を発揮可能なことを示す.

  本研究の構成は以下である.
  \secref{sec:proposed}では, ワイヤ配置の設計パラメータの設定, 目的関数の設定, 多目的最適化について順に述べる.
  \secref{sec:experiment}では, 2関節3リンクの平面マニピュレータについて, 重力の影響を受ける場合と受けない場合, モーメントアームが一定な場合と可変な場合, ワイヤ本数やワイヤ経由点の個数が異なる場合について, 多様なワイヤ配置を構築する.
  \secref{sec:discussion}では実験結果について考察し, \secref{sec:conclusion}で結論を述べる.
}%

\section{Design Optimization of Wire Arrangement with Variable Relay Points} \label{sec:proposed}
\switchlanguage%
{%
  First, we set discrete and continuous parameters for wire arrangement.
  Next, we define target operational force and velocity spaces, along with their corresponding objective functions.
  Lastly, we conduct multi-objective black-box optimization based on these design parameters and objectives.
}%
{%
  まずワイヤ配置設計の離散値・連続値パラメータを設定する.
  次に目標手先力空間・目標手先速度空間を設定し, これを実現するための評価関数の設定を行う.
  最後にこれら評価値に基づく多目的最適化ブラックボックスを行う方法を紹介する.
}%

\begin{figure}[t]
  \centering
  \includegraphics[width=0.85\columnwidth]{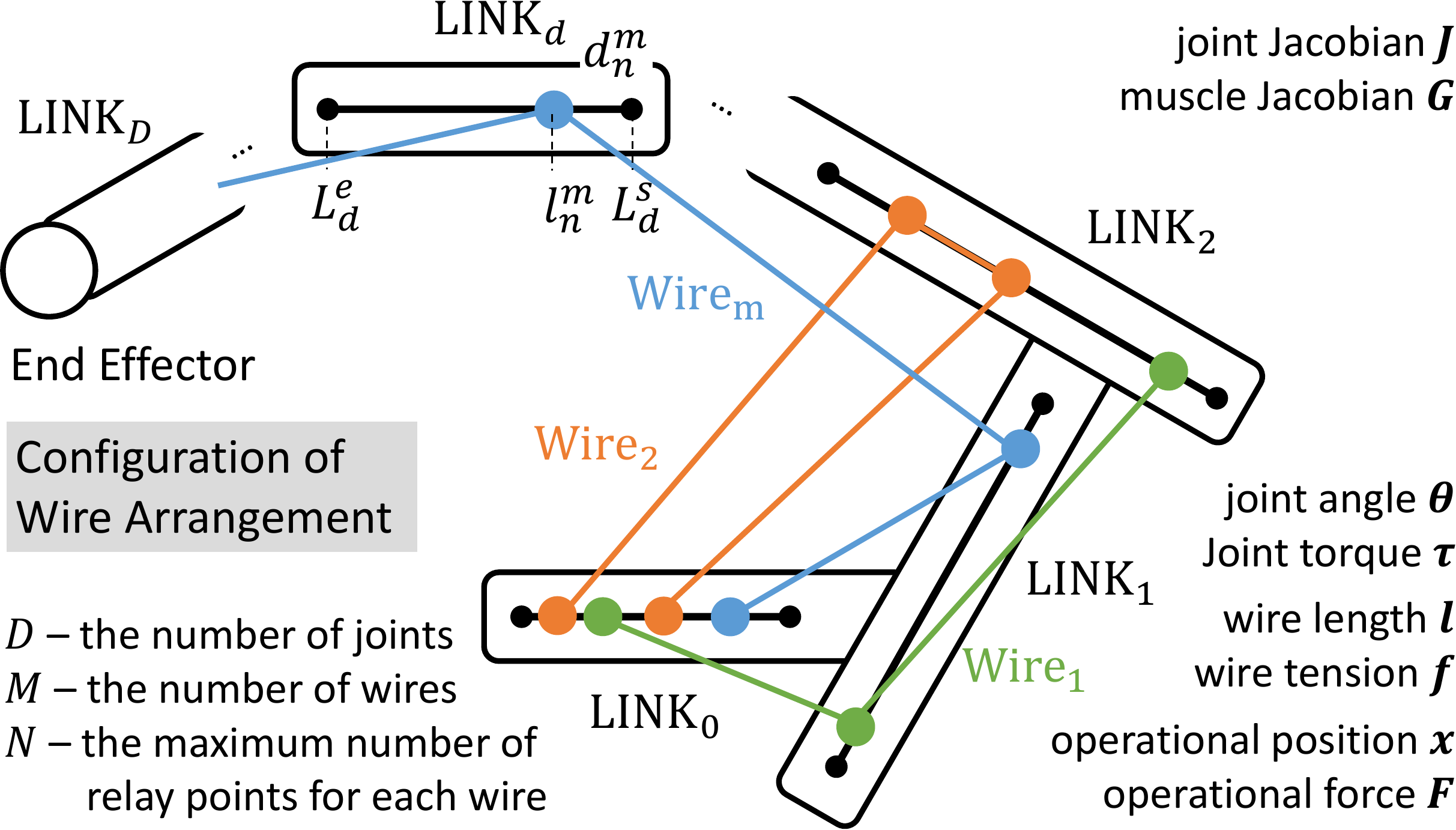}
  \vspace{-1.0ex}
  \caption{The design parameters of wire arrangement. The number of wires $M$, the number of relay points $N$, the position of the relay point $l^m_n$, and the link $d^m_n$ that the relay point is attached to are set as variables.}
  \label{figure:design-params}
  \vspace{-3.0ex}
\end{figure}

\subsection{Design Parameters of Wire Arrangement} \label{subsec:design-params}
\switchlanguage%
{%
  In this study, the joint structure is predefined and only the wire arrangement is optimized.
  All motors and the start points of wires are located at the root of the robot.
  This structure maximizes the advantage of tendon-driven robots, in that the weight of the movable part can be reduced by separating the actuators and links.
  Note that the problem setting of this study is feasible in hardware, exemplified by the legs of the kangaroo robot \cite{yoshimura2023kangaroo} shown in \figref{figure:concept}.

  An overview of the design parameters is shown in \figref{figure:design-params}.
  We consider a robot with $M$ wires for a given link structure with $D$ joints.
  Let $\textrm{LINK}_{0}$ be the first link that the actuators are attached to, and $\textrm{LINK}_{D}$ be the last link.
  We set the end of $\textrm{LINK}_{D}$ as an end effector.
  Let $N$ ($N \geq 2$) be the maximum number of relay points (including start and end points) for each wire.
  The $n$-th ($1 \leq n \leq N$) relay point on the $m$-th ($1 \leq m \leq M$) wire is attached to the link $d^{m}_{n}$ ($0 \leq d^{m}_{n} \leq D$).
  The position of the relay point attached to the link is expressed as $l^{m}_{n}$ ($0.0 \leq l^{m}_{n} \leq 1.0$).
  For simplicity, the relay points of $\textrm{LINK}_{d}$ ($0 \leq d \leq D$) are arranged in a straight line on the link, and the position is within $[L^{s}_{d}, L^{e}_{d}]$ ($L^{\{s, e\}}_{d}$ is a constant).
  That is, on $\textrm{LINK}_{d}$, $l^{m}_{n}=0$ represents the position $L^{s}_{d}$, $l^{m}_{n}=1$ represents the position $L^{e}_{d}$, and the actual link position is $L^{s}_{d}+l^{m}_{n}(L^{e}_{d}-L^{s}_{d})$.
  The first relay point can be attached to only $\textrm{LINK}_{0}$, while the rest of the relay points can be attached to $LINK_{d}$ ($0 \leq d \leq D$).
  In summary, the design parameter of each wire $m$ is a continuous value of $l^{m}_{1}$ at $n=1$, and a continuous value of $l^{m}_{n}$ and a discrete value of $d^{m}_{n}$ with $D+1$ choices at $n \geq{2}$.

  As a comparison, we also perform experiments for the case with constant moment arm.
  The design parameters of each wire $m$ are represented by continuous values of moment arm $r^{m}_{d}$ ($0.0 \leq r^{m}_{d} \leq 1.0$) for each joint $d$.
  Let $[R^{s}_{d}, R^{e}_{d}]$ denote the range of possible moment arms at each joint $d$, $r^{m}_{d}=0$ represents the radius $R^{s}_{d}$, $r^{m}_{d}=1$ represents the radius $R^{e}_{d}$, and the actual moment arm is $R^{s}_{d}+r^{m}_{d}(R^{e}_{d}-R^{s}_{d})$.
  As described in \secref{sec:introduction}, while this configuration has high performance because the moment arm can be designed directly, the ease of wire arrangement is lost and the large moment arm becomes difficult to obtain.
}%
{%
  本研究では関節構造は予め定めておき, ワイヤ配置のみの最適化を行う.
  また, 全てのモータはアームの根本に配置されており, ワイヤの始点も同様にアームの根本である.
  これは, ワイヤ駆動系ロボットにおける, アクチェータと可動部を分離することで可動部の重量を減らすことが可能な利点を最大限活かした構造と言える.
  なお, 本研究の実験設定は実際にハードウェアとして実現可能であり, \figref{figure:concept}に示すカンガルーロボット\cite{yoshimura2023kangaroo}の脚はその例である.

  設計パラメータの概要を\figref{figure:design-params}に示す.
  本研究では, 与えられた$D$個の関節を持つリンク構造に対して, $M$本のワイヤを張ったロボットを考える.
  最も根本のワイヤの始点となるリンクを$\textrm{LINK}_{0}$とし, 末端のリンクが$\textrm{LINK}_{D}$となる.
  $\textrm{LINK}_{D}$の末端をエンドエフェクタとする.
  各ワイヤの経由点(これは始点と終止点を含むこととする)の最大数を$N$ ($N\geq2$)とする.
  $m$ ($1\leq{m}\leq{M}$)本目のワイヤにおける$n$ ($1\leq{n}\leq{N}$)番目の経由点について, 経由点を取り付けるリンク$d^{m}_{n}$ ($0\leq{d}^{m}_{n}\leq{D}$)を指定する.
  そして, 指定したリンクに対する経由点の取り付け位置を$l^{m}_{n}$ ($0 \leq l^{m}_{n} \leq 1$)と表現する.
  $\textrm{LINK}_{d}$ ($0\leq{d}\leq{D}$)に対して取り付け可能な経由点の位置は, 簡単のためそのリンクの直線上とし, $\textrm{LINK}_{d}$における取付可能な始点位置を$L^{s}_{d}$, 終点位置を$L^{e}_{d}$とする.
  つまり, $\textrm{LINK}_{d}$上において, $l^{m}_{n}=0$は位置$L^{s}_{d}$, $l^{m}_{n}=1$は位置$L^{e}_{d}$を表し, 実際のリンク位置は$L^{s}_{d}+l^{m}_{n}(L^{e}_{d}-L^{s}_{d})$となる.
  1つ目の経由点が取り付け可能なリンクは$\textrm{LINK}_{0}$のみであるが, 2つ目以降の経由点が取り付け可能なリンクは$LINK_{d}$ ($0\leq{d}\leq{D}$)のいずれかとなる.
  まとめると, 各ワイヤ$m$の設計パラメータは$n=1$の経由点において$l^{m}_{1}$の連続値, $n (\geq{2})$の経由点において$l^{m}_{n}$の連続値と$D+1$個の選択肢を有する離散値$d^{m}_{n}$となる.

  なお, 比較としてモーメントアーム一定の場合の実験も行う.
  この場合, 各ワイヤ$m$の設計パラメータは, 各関節に対するモーメントアームの大きさ$r^{m}_{d}$ ($0\leq{r}^{m}_{d}\leq{1}$)の連続値によって表現される.
  各関節$d$における可能なモーメントアームの範囲を$[R^{s}_{d}, R^{e}_{d}]$とすると, $r^{m}_{d}=0$は位置$R^{s}_{d}$, $r^{m}_{d}=1$は位置$R^{e}_{d}$を表し, 実際のモーメントアームは$R^{s}_{d}+r^{m}_{d}(R^{e}_{d}-R^{s}_{d})$となる.
  \secref{sec:introduction}でも述べたように, この構成はモーメントアームを直接自由に設計でき高い性能を有する反面, ワイヤ配置の容易さが失われ, かつモーメントアームも大きく取りにくいという欠点がある.
}%

\begin{figure}[t]
  \centering
  \includegraphics[width=0.6\columnwidth]{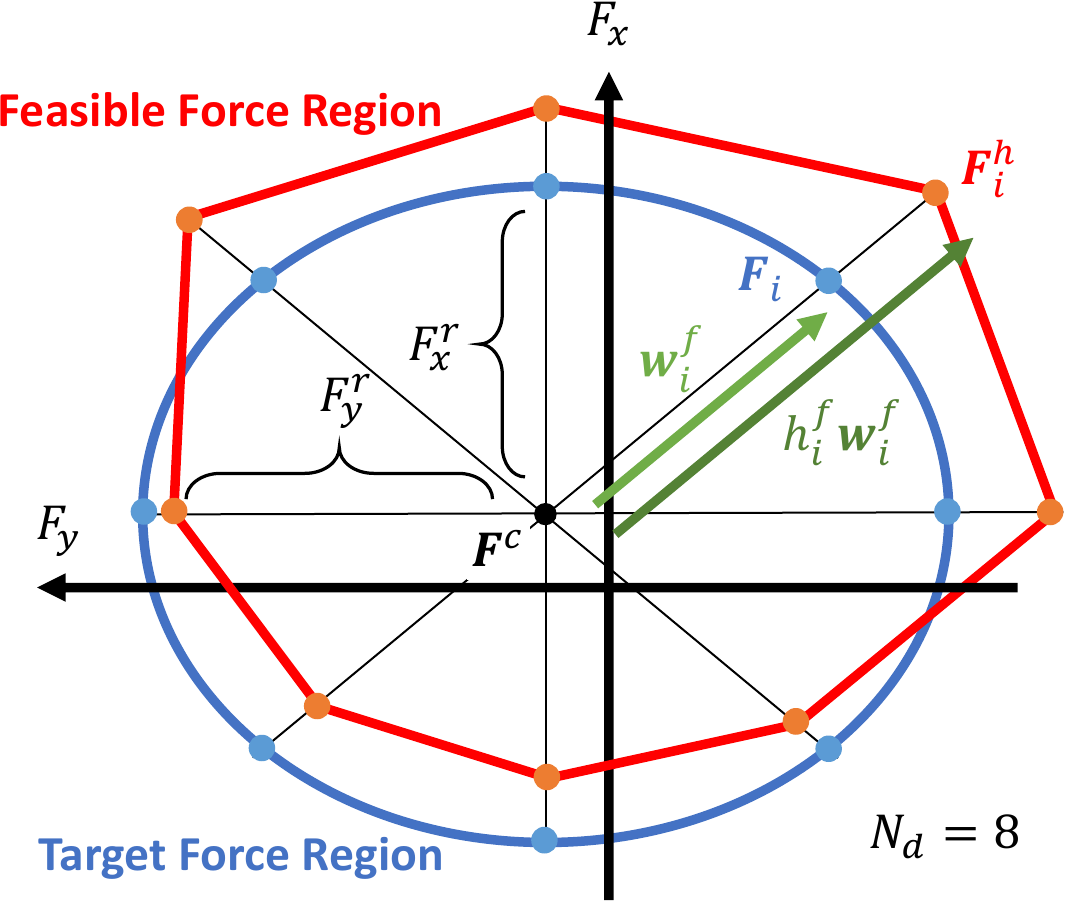}
  \vspace{-1.0ex}
  \caption{The definition of variables for calculation of objective function for feasible operational force space.}
  \label{figure:design-obj}
  \vspace{-3.0ex}
\end{figure}

\subsection{Calculation of Objective Functions} \label{subsec:design-obj}
\switchlanguage%
{%
  Objective functions are calculated for the obtained designs.
  In this study, we use feasible operational force space (OFS) and operational velocity space (OVS) as the objective functions.
  We define the target OFS and OVS, and find design solutions that realize them as much as possible.
  We consider OFS and OVS in a two-dimensional (2D) plane, but this can be extended to three dimensions (3D) in a similar manner.

  First, some basic formulas are described.
  Let $\bm{f}$ be wire tension, $\bm{\tau}$ be joint torque, and $\bm{F}$ be operational force of the end effector.
  Also, let $\bm{l}$ be wire length, $\bm{\theta}$ be joint angle, and $\bm{x}$ be operational position of the end effector.
  Here, the following relations generally hold,
  \begin{align}
    \bm{l} &= \bm{g}_{m}(\bm{\theta}) \\
    \dot{\bm{l}} &= \bm{G}(\bm{\theta})\dot{\bm{\theta}} \\
    \bm{\tau} &= -\bm{G}^{T}(\bm{\theta})\bm{f} \\
    \bm{x} &= \bm{g}_{j}(\bm{\theta}) \\
    \dot{\bm{x}} &= \bm{J}(\bm{\theta})\dot{\bm{\theta}} \\
    \bm{\tau} &= \bm{J}^{T}(\bm{\theta})\bm{F}
  \end{align}
  where $\bm{g}_{m}$ denotes the mapping from $\bm{\theta}$ to $\bm{l}$ and $\bm{g}_{j}$ denotes the mapping from $\bm{\theta}$ to $\bm{x}$.
  Also, $\bm{G}$ denotes the muscle Jacobian and $\bm{J}$ denotes the joint Jacobian.
  The minimum and maximum values of wire tension $\bm{f}$ are denoted by $\bm{f}^{min}$ and $\bm{f}^{max}$, and those of wire velocity $\dot{\bm{l}}$ are denoted by $\dot{\bm{l}}^{min}$ and $\dot{\bm{l}}^{max}$.
  In this study, we do not change the ranges dynamically by considering back electromotive force.

  Next, we describe the objective function for the feasible OFS (\figref{figure:design-obj}).
  Defining the target OFS in a 2D plane as an ellipse, its parameters are the center point of the ellipse $\bm{F}^{c} = \begin{pmatrix}F^{c}_{x} & F^{c}_{y}\end{pmatrix}^{T}$ and the radius of the ellipse for the $x$- and $y$-axes $F^{r}_{\{x, y\}}$.
  Note that the feasible OFS and OVS are always convex.
  When this ellipse is divided into $N_{d}$ pieces, the operational force $\bm{F}_{i}$ at each point $i$ ($0 \leq i < N_{d}$) on the ellipse can be expressed as follows.
  \begin{align}
    \bm{w}^{f}_{i} &:= \begin{pmatrix}
      F^{r}_{x}\cos(2\pi{i}/N_{d})\\
      F^{r}_{y}\sin(2\pi{i}/N_{d})
    \end{pmatrix}\\
    \bm{F}_{i} &:= \bm{w}^{f}_{i} + \bm{F}^{c}
  \end{align}
  Here, for each point $i$, we introduce a value $h^{f}_{i}$ that expresses by how much the feasible OFS exceeds the target OFS, and define the following point $\bm{F}^{h}_{i}$.
  \begin{align}
    \bm{F}^{h}_{i} := h^{f}_{i}\bm{w}^{f}_{i} + \bm{F}^{c}
  \end{align}
  We maximize $h^{f}_{i}$ by the following linear programming.
  \begin{align}
    \underset{h^{f}_{i}, \bm{f}}{\textrm{maximize}} & \;\;\;\;\;\;\;\;\;\;\;\;\;\;\;\;\;\; h^{f}_{i} \label{eq:f-max}\\
    \textrm{subject to} &\;\; -\bm{G}^{T}(\bm{\theta})\bm{f} = \bm{J}^{T}(\bm{\theta})\bm{F}^{h}_{i} \label{eq:f-st-1}\\
    &\;\;\;\;\; \bm{f}^{min} \leq \bm{f} \leq \bm{f}^{max} \label{eq:f-st-2}
  \end{align}
  When $h^{f}_{i} = 1$, the target and feasible OFS coincide for the direction $\bm{w}^{f}_{i}$, and $h^{f}_{i} \geq 1$ should be satisfied.
  Therefore, we compute $h^{f}_{i}$ for each point $i$, set the following $E_{force}$ as the objective function, and minimize it.
  \begin{align}
    E_{force} := \sum_{i}{\max(1-h^{f}_{i}, 0)}
  \end{align}
  If gravity compensation is considered, $\bm{F}^{c}$ should be obtained by the following quadratic programming,
  \begin{align}
    \underset{\bm{F}^{c}}{\textrm{minimize}} & \;\;\;\;\;\;\;\;\;\; ||\bm{F}^{c}||^{2}_{2}\\
    \textrm{subject to} &\;\;\;\;\;\; \bm{\tau}_{g} = \bm{J}^{T}(\bm{\theta})F^{c}
  \end{align}
  where $\bm{\tau}_{g}$ is the gravity compensation torque required at the current joint angle $\bm{\theta}$ and $||\cdot||_{2}$ is the L2 norm.
  Note that in practice, since $\bm{J}^{T}(\bm{\theta})\bm{F}^{c} = \bm{\tau}_{g}$ in \equref{eq:f-st-1}, the calculation of $\bm{F}_{c}$ is not necessary for optimization, but only for drawing the elliptic center.

  Next, we describe the objective function for the feasible OVS.
  Defining the target OVS in a 2D plane as an ellipse, its parameters are the radius $v^{r}_{\{x, y\}}$ of the ellipse for the $x$- and $y$-axes.
  When this ellipse is divided into $N_{d}$ pieces, the operational velocity $\bm{v}_{i}$ at each point $i$ ($0 \leq i < N_{d}$) on the ellipse can be expressed as follows.
  \begin{align}
    \bm{w}^{v}_{i} &:= \begin{pmatrix}
      v^{r}_{x}\cos(2\pi{i}/N_{d})\\
      v^{r}_{y}\sin(2\pi{i}/N_{d})
    \end{pmatrix}\\
    \bm{v}_{i} &:= \bm{w}^{v}_{i}
  \end{align}
  Here, for each point $i$, we introduce a value $h^{v}_{i}$ that expresses by how much the feasible OVS exceeds the target OVS, and define the following point $\bm{v}^{h}_{i}$.
  \begin{align}
    \bm{v}^{h}_{i} := h^{v}_{i}\bm{w}^{v}_{i}
  \end{align}
  We maximize $h^{v}_{i}$ by the following linear programming.
  \begin{align}
    \underset{h^{v}_i, \dot{\bm{\theta}}}{\textrm{maximize}} & \;\;\;\;\;\;\;\;\;\;\;\;\;\;\;\;\;\; h^{v}_i \label{eq:v-max}\\
    \textrm{subject to} &\;\;\;\;\;\; \bm{J}(\bm{\theta})\dot{\bm{\theta}} = \bm{v}^{h}_{i} \label{eq:v-st-1}\\
    &\;\;\;\;\; \dot{\bm{l}}^{min} \leq \bm{G}(\bm{\theta})\dot{\bm{\theta}} \leq \dot{\bm{l}}^{max} \label{eq:v-st-2}
  \end{align}
  Similarly to $E_{force}$, we set the following $E_{velocity}$ as the objective function, and minimize it.
  \begin{align}
    E_{velocity} := \sum_{i}{\max(1-h^{v}_{i}, 0)}
  \end{align}
  Note that we can define target spaces of various shapes other than ellipsoids, and can freely define various objective functions such as operational wrench force or rotation velocity.
}%
{%
  得られた設計に対して評価値を計算する.
  本研究では, 評価値として発揮可能手先力空間・発揮可能手先速度空間を用いる.
  目標手先力空間・目標手先速度空間を定義し, これをできる限り実現する設計解を見つけることを行う.
  ここでは二次元平面における手先力・手先速度空間について考えるが, 同様の形で3次元に拡張することが可能である.

  まずいくつか前提について述べる.
  本研究ではロボットが発揮するワイヤ張力を$\bm{f}$, 関節トルクを$\bm{\tau}$, 手先力を$\bm{F}$とする.
  また, ワイヤ長さを$\bm{l}$, 関節角度を$\bm{\theta}$, 手先位置を$\bm{x}$とする.
  このとき, 一般的に以下の関係が成り立つ.
  \begin{align}
    \bm{l} &= \bm{g}_{m}(\bm{\theta}) \\
    \dot{\bm{l}} &= \bm{G}(\bm{\theta})\dot{\bm{\theta}} \\
    \bm{\tau} &= -\bm{G}^{T}(\bm{\theta})\bm{f} \\
    \bm{x} &= \bm{g}_{j}(\bm{\theta}) \\
    \dot{\bm{x}} &= \bm{J}(\bm{\theta})\dot{\bm{\theta}} \\
    \bm{\tau} &= \bm{J}^{T}(\bm{\theta})\bm{F}
  \end{align}
  ここで, $\bm{g}_{m}$は$\bm{\theta}$から$\bm{l}$への写像, $\bm{g}_{j}$は$\bm{\theta}$から$\bm{x}$への写像を表す.
  また, $\bm{G}$は筋長ヤコビアン, $\bm{J}$は関節ヤコビアンを表す.
  ワイヤ張力$\bm{f}$の最小値を$\bm{f}^{min}$, 最大値を$\bm{f}^{max}$, ワイヤ速度$\dot{\bm{l}}$の最小値を$\dot{\bm{l}}^{min}$, 最大値を$\dot{\bm{l}}^{max}$とする.
  なお, 本研究では逆起電力等を考え動的に定義範囲を変更することは行わない.

  次に, 発揮可能手先力空間の評価関数について述べる(\figref{figure:design-obj}).
  二次元平面における理想的な手先力空間を楕円として定義すると, そのパラメータは楕円の中心点$\bm{F}^{c} = \begin{pmatrix}F^{c}_{x} & F^{c}_{y}\end{pmatrix}^{T}$, $x$軸方向と$y$軸方向に関する楕円の半径$F^{r}_{\{x, y\}}$によって表される.
  なお, 発揮可能手先力空間・手先速度空間は必ず凸になる.
  この楕円を$N_{d}$個に分割したとき, 楕円上の各点$i$ ($0 \leq i < N_{d}$)における手先力$\bm{F}_{i}$は以下のように表現できる.
  \begin{align}
    \bm{w}^{f}_{i} &:= \begin{pmatrix}
      F^{r}_{x}\cos(2\pi{i}/N_{d})\\
      F^{r}_{y}\sin(2\pi{i}/N_{d})
    \end{pmatrix}\\
    \bm{F}_{i} &:= \bm{w}^{f}_{i} + \bm{F}^{c}
  \end{align}
  ここで, 各点$i$について, どの程度その身体の発揮可能手先力空間が目標手先力空間を上回っているかを表現する値$h^{f}_{i}$を導入し, 以下の点$\bm{F}^{h}_{i}$を定義する.
  \begin{align}
    \bm{F}^{h}_{i} := h^{f}_{i}\bm{w}^{f}_{i} + \bm{F}^{c}
  \end{align}
  これらから, 以下の線形計画問題を定義する.
  \begin{align}
    \underset{h^{f}_{i}, \bm{f}}{\textrm{maximize}} & \;\;\;\;\;\;\;\;\;\;\;\;\;\;\;\;\;\; h^{f}_{i} \label{eq:f-max}\\
    \textrm{subject to} &\;\; -\bm{G}^{T}(\bm{\theta})\bm{f} = \bm{J}^{T}(\bm{\theta})\bm{F}^{h}_{i} \label{eq:f-st-1}\\
    &\;\;\;\;\; \bm{f}^{min} \leq \bm{f} \leq \bm{f}^{max} \label{eq:f-st-2}
  \end{align}
  得られた$h^{f}_{i}$が$1$であるとき, 方向$\bm{w}^{f}_{i}$について目標手先力空間と発揮可能手先力空間が一致することを表し, $h^{f}_{i}\geq{1}$であることが望ましい.
  よって, これを各点$i$について計算し, 以下の$E_{force}$を評価関数とし, これを最小化する.
  \begin{align}
    E_{force} := \sum_{i}{\max(1-h^{f}_{i}, 0)}
  \end{align}
  もし重力補償を考える場合, $\bm{F}^{c}$は以下の二次計画法を用いて求める必要がある.
  \begin{align}
    \underset{\bm{F}^{c}}{\textrm{minimize}} & \;\;\;\;\;\;\;\;\;\; ||\bm{F}^{c}||^{2}_{2}\\
    \textrm{subject to} &\;\;\;\;\;\; \bm{\tau}_{g} = \bm{J}^{T}(\bm{\theta})F^{c}
  \end{align}
  ここで, $\bm{\tau}_{g}$は現在の関節角度$\bm{\theta}$において必要な重力補償トルクであり, $||\cdot||_{2}$はL2ノルムを表す.
  なお, 実際には\equref{eq:f-st-1}において $\bm{J}^{T}(\bm{\theta})\bm{F}^{c} = \bm{\tau}_{g}$となるため, この$\bm{F}_{c}$の計算は最適化に必要なく, 楕円中心の描画にしか必要ない.

  次に, 発揮可能手先速度空間の評価関数について述べる.
  二次元平面における理想的な手先速度空間を同様に楕円として定義すると, そのパラメータは$x$軸方向と$y$軸方向に関する楕円の半径$v^{r}_{\{x, y\}}$によって表される.
  この楕円を$N_{d}$個に分割したとき, 楕円上の各点$i$ ($0 \leq i < N_{d}$)における手先速度$\bm{v}_{i}$は以下のように表現できる.
  \begin{align}
    \bm{w}^{v}_{i} &:= \begin{pmatrix}
      v^{r}_{x}\cos(2\pi{i}/N_{d})\\
      v^{r}_{y}\sin(2\pi{i}/N_{d})
    \end{pmatrix}\\
    \bm{v}_{i} &:= \bm{w}^{v}_{i}
  \end{align}
  ここで, 各点$i$について, どの程度その身体の発揮可能手先速度空間が目標手先速度空間を上回っているかを表現する値$h^{v}_{i}$を導入し, 以下の点$\bm{v}^{h}_{i}$を定義する.
  \begin{align}
    \bm{v}^{h}_{i} := h^{v}_{i}\bm{w}^{v}_{i}
  \end{align}
  これらから, 以下の線形計画問題を定義する.
  \begin{align}
    \underset{h^{v}_i, \dot{\bm{\theta}}}{\textrm{maximize}} & \;\;\;\;\;\;\;\;\;\;\;\;\;\;\;\;\;\; h^{v}_i \label{eq:v-max}\\
    \textrm{subject to} &\;\;\;\;\;\; \bm{J}(\bm{\theta})\dot{\bm{\theta}} = \bm{v}^{h}_{i} \label{eq:v-st-1}\\
    &\;\;\;\;\; \dot{\bm{l}}^{min} \leq \bm{G}(\bm{\theta})\dot{\bm{\theta}} \leq \dot{\bm{l}}^{max} \label{eq:v-st-2}
  \end{align}
  同様に以下の$E_{velocity}$を評価関数とし, これを最小化する.
  \begin{align}
    E_{velocity} := \sum_{i}{\max(1-h^{v}_{i}, 0)}
  \end{align}
  もちろん楕円体以外にも多様な形の目標空間を定義できるうえ, 手先トルクや手先回転速度など, 自由に評価指標を設定できる.
}%

\subsection{Design of Wire Arrangement Using Multi-Objective Black-Box Optimization} \label{subsec:blackbox-optimization}
\switchlanguage%
{%
  We have formulated the design parameters and the objective functions.
  By using them, multi-objective optimization is performed to determine the design parameters.
  In this study, we use NSGA-II, a multi-objective optimization method implemented in optuna \cite{akiba2019optuna}, a black box optimization library.
  NSGA-II was chosen for its ability to perform multi-objective optimization, handle both continuous and discrete parameters, and handle a relatively large number of samples.
  Note that if no solution is found for the linear programming problem in \equref{eq:f-max}--\equref{eq:f-st-2} or \equref{eq:v-max}--\equref{eq:v-st-2}, the design is pruned.
  In the experiments, we illustrate the obtained Pareto solutions with $N_{sample}$ as the number of samples, and discuss the wire arrangement and the feasible OFS and OVS of the design.
  Although only two objective functions are used, multi-objective optimization is also possible by using other objective functions such as minimizing the wire length to suppress elongation or adding constraints on the number and positions of relay points.
}%
{%
  これまでにワイヤ配置設計パラメータとその評価指標$E_{\{force, velocity\}}$を得た.
  これらを用いて多目的最適化を行い, 身体パラメータを決定する.
  本研究ではブラックボックス最適化のライブラリであるoptuna \cite{akiba2019optuna}に実装された多目的最適化手法であるNSGA-IIを用いる.
  NSGA-IIは, 多目的最適化ができること, 連続値と離散値の両方を最適化パラメータとして利用できること, 推奨されるサンプリング数が比較的大きいことから選択した.
  なお, \equref{eq:f-max}--\equref{eq:f-st-2}や\equref{eq:v-max}--\equref{eq:v-st-2}における線形計画問題の解が見つからない場合, その設計については枝刈りを行う.
  サンプリング回数を$N_{sample}$として, 得られたパレート解を図示し, 実際のワイヤ配置と各解における発揮可能手先力・手先速度空間について考察する.
  本研究では2つの評価関数のみ用いているが, この他に, ワイヤ長を最小化して伸びを抑制したり, 経由点の個数や位置に関する制約などを導入した多目的最適化も可能である.
}%

\begin{figure*}[t]
  \centering
  \includegraphics[width=1.7\columnwidth]{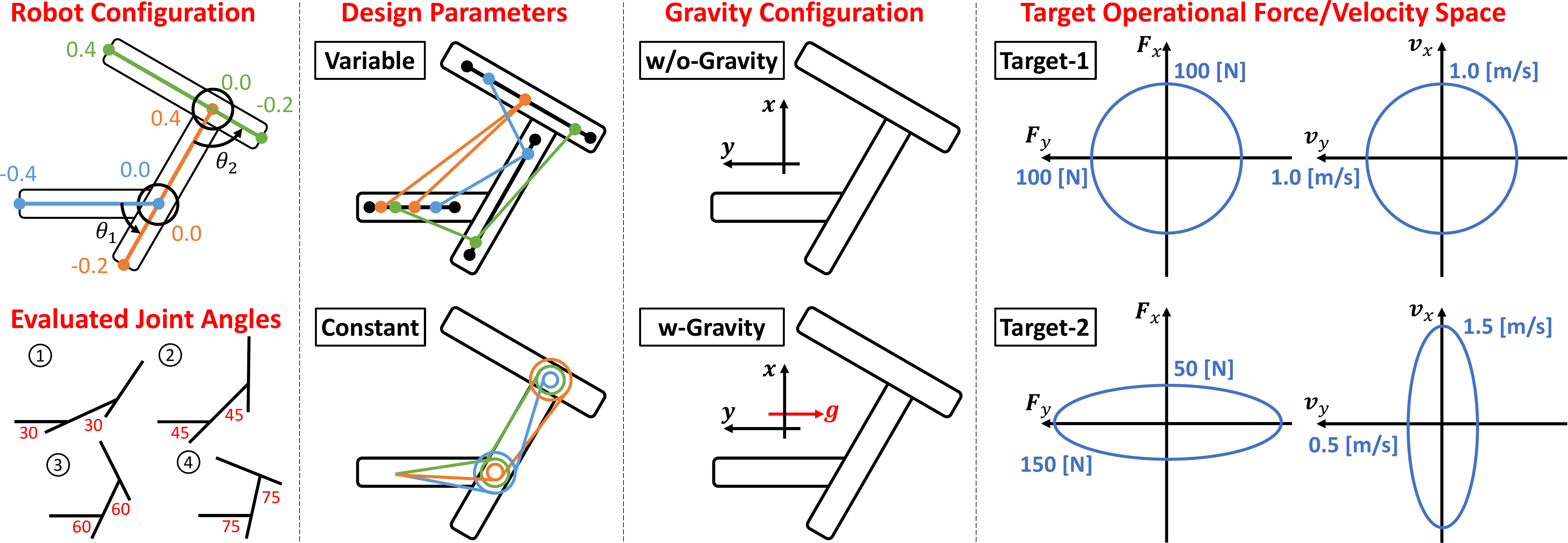}
  \vspace{-1.0ex}
  \caption{The experimental setup of this study. The robot configuration for $L^{\{s, e\}}_{d}$ and the evaluated joint angles are shown. The design parameters were changed to \textbf{Variable} or \textbf{Constant}, gravity configuration to \textbf{w/o-Gravity} or \textbf{w-Gravity}, and target operational force / velocity space to \textbf{Target-1} or \textbf{Target-2}.}
  \label{figure:exp-setup}
  \vspace{-3.0ex}
\end{figure*}

\section{Experiments} \label{sec:experiment}
\switchlanguage%
{%
  The experimental setup is shown in \figref{figure:exp-setup}.
  We focus on a two-joint and three-link mechanism in a 2D plane for detailed experiments and comparisons.
  The robot model used in this study is shown under ``Robot Configuration'' in \figref{figure:exp-setup}, where $\textrm{LINK}_{0}$ has a length of 0.4 m and $\textrm{LINK}_{\{1, 2\}}$ has a length of 0.6 m.
  Relay points can be attached to any of three links from one end to the other.
  $\textrm{LINK}_{0}$ is fixed and the weight of $\textrm{LINK}_{\{1, 2\}}$ is set to 4 kg assuming the density of aluminum.
  The total value of $E_{\{force, velocity\}}$ for the four joint angles (\ctext{1}--\ctext{4}) is used as the actual objective function, where the angle of each joint is changed by 15 deg as shown under ``Evaluated Joint Angles'' in \figref{figure:exp-setup}.
  Note that we have chosen this configuration for reasons of computational complexity, visualization, and the desire to emphasize significant changes in joint Jacobian due to the change in $\bm{\theta}_{1}+\bm{\theta}_{2}$, though a wider range of joint angle states should be considered.
  As a comparison, several experiments are conducted by changing the design parameters and settings.
  First, as shown under ``Design Parameters'' in \figref{figure:exp-setup}, we conduct experiments for two types: \textbf{Variable} in which each relay point can be freely selected, and \textbf{Constant} in which the muscle Jacobian is kept constant using pulleys.
  In this comparison, we vary the number of wires $M$ and the maximum number of relay points $N$.
  For \textbf{Constant}, the range of moment arms of the wire for each joint is set to [$R^{s}_{d}$, $R^{e}_{d}$]=[-0.1, 0.1] [m] (since the distance between the joints is 0.4 m, the moment arm can be increased to 0.2, but we set it to 0.1 based on the feasibility. In some experiments, we set [-0.4, 0.4] [m] in order to investigate the maximum possible performance).
  We also compare two types: \textbf{w/o-Gravity} which is not affected by gravity, and \textbf{w/-Gravity} which is affected by gravity as in the human arm as shown under ``Gravity Configuration''.
  In addition, as shown under ``Target Operational Force/Velocity Space'', we perform experiments by setting types of target OFS and OVS: \textbf{Target-1} and \textbf{Target-2}.
  For other parameters, we set $f^{min}=10$ [N], $f^{max}=200$ [N], $\dot{l}^{min}=-0.4$ [m/s], $\dot{l}^{max}=0.4$ [m/s], $N_{d}=8$, and $N_{sample}=10000$ (the performance is equivalent to that of a Maxon 90W BLDC motor with 29:1 gear ratio).
  For each experiment, the sampling results and Pareto solutions are shown, and the target space (blue lines) and the feasible space (red lines) of force and velocity are shown for the design with the smallest $|E_{force}-E_{velocity}|$ (the solution in which $E_{force}$ and $E_{velocity}$ are considered equally).
  For \textbf{Variable}, its wire arrangement is shown as a figure, and for \textbf{Constant}, the moment arm of each wire for each joint is shown since the wire arrangement is difficult to visualize.
}%
{%
  本研究の実験セットアップを\figref{figure:exp-setup}に示す.
  本研究では2次元平面における2関節3リンク機構に絞って詳細な実験・比較を行う.
  用いるロボットモデルは\figref{figure:exp-setup}における``Robot Configuration''のようになっており, $\textrm{LINK}_{0}$は長さ0.4 m, $\textrm{LINK}_{\{1, 2\}}$は長さ0.6 mである.
  リンクの端から端までのいずれにも経由点を取り付け可能なパラメータ設定となっている.
  また, $\textrm{LINK}_{0}$は固定されており, $\textrm{LINK}_{\{1, 2\}}$の重さはアルミの密度を仮定し4 kgと設定している.
  本研究で評価値を計算する際は, \figref{figure:exp-setup}の``Evaluated Joint Angles''に示す各関節の角度を15 degずつ変化させた4種類の関節角度状態(\ctext{1}--\ctext{4})における$E_{\{force, velocity\}}$の合計値を用いている.
  なお, 本来はより多様な関節角度状態を扱うべきであるが, 計算量や図示の観点, また, 関節ヤコビアンの変化に大きく寄与する$\bm{\theta}_{1}+\bm{\theta}_{2}$をなるべく大きく変化させたいという観点からこのような設定としている.
  本研究では比較として, 設計パラメータや設定を変更して複数の実験を行っている.
  まず, \figref{figure:exp-setup}の``Design Parameters''に示すように, ワイヤ経由点を自由に選ぶことができるパターン\textbf{Variable}と, 筋長ヤコビアンが一定なパターン\textbf{Constant}の2種類について実験を行う.
  なお, この際ワイヤ数$M$やワイヤ経由点の最大数$N$を変化させて比較を行う.
  \textbf{Constant}については, ワイヤの各関節に対するモーメントアームの範囲を[$R^{s}_{d}$, $R^{e}_{d}$]=[-0.1, 0.1] [m]としている(関節間の距離が0.4 mであるため, 最大0.2までモーメントアームを取ることができるが, 実現性を踏まえ0.1とした, 一部の実験では可能な最大のパフォーマンスを調べるために[-0.4, 0.4] [m]としている).
  また, ``Gravity Configuration''に示すように重力の影響を受けないスカラ型のようなパターン(\textbf{w/o-Gravity}), 人間の肩甲骨-上腕-前腕のように重力の影響を受けるパターン(\textbf{w/-Gravity})の2種類についても比較を行う.
  加えて, ``Target Operational Force/Velocity Space''に示すように, 目標手先力空間と目標手先速度空間を\textbf{Target-1}と\textbf{Target-2}の二種類に変化させて実験を行う.
  その他のパラメータについては, $f^{min}=10$ [N], $f^{max}=200$ [N], $\dot{l}^{min}=-0.4$ [m/s], $\dot{l}^{max}=0.4$ [m/s], $N_{d}=8$, $N_{sample}=10000$としている(Maxon 90W BLDC motor with 29:1 gear ratioと同等の性能を採用した).
  それぞれの実験について, サンプリング結果とパレート解を示し, そのうち$|E_{force}-E_{velocity}|$が最も小さな設計解($E_{force}$と$E_{velocity}$を同じくらい考慮した解)について, 目標値(青線)と発揮可能手先力・手先速度空間(赤線)を図示する.
  また, \textbf{Variable}については, そのワイヤ配置を図示し, \textbf{Constant}については, ワイヤ配置の描画が困難なためワイヤの各関節に対するモーメントアームを記載する.
}%

\begin{figure*}[t]
  \centering
  \includegraphics[width=1.85\columnwidth]{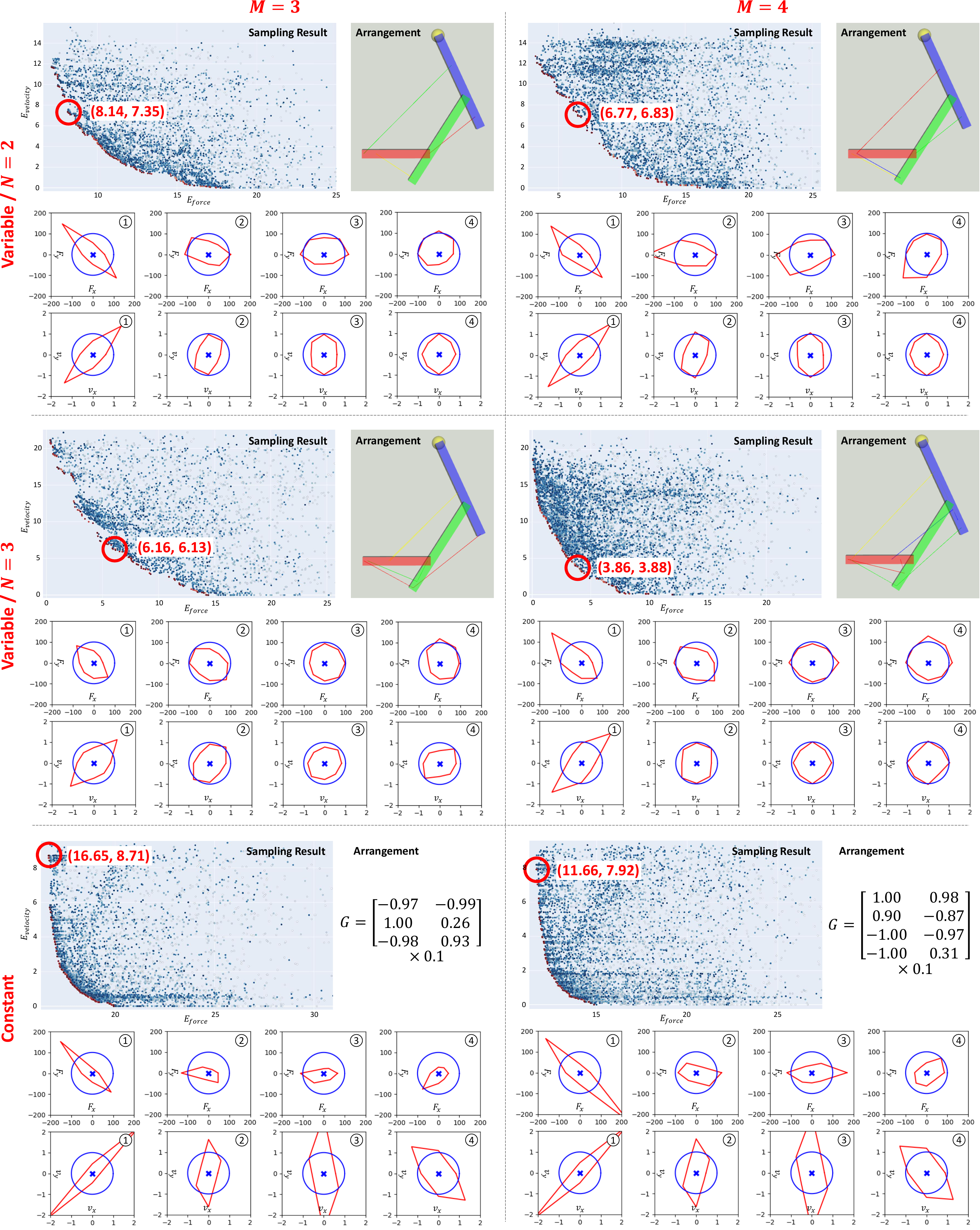}
  \vspace{-1.0ex}
  \caption{The experiment of Target-1 w/o Gravity. For \textbf{Variable} with $M=\{3, 4\}$ and $N=\{2, 3\}$ and for \textbf{Constant} with $M=\{3, 4\}$, the sampling results and Pareto solutions are shown. For one Pareto solution with minimized $|E_{force}-E_{velocity}|$, the wire arrangement and target (blue line) and feasible (red line) operational force / velocity spaces are shown.}
  \label{figure:exp1-pareto}
  \vspace{-3.0ex}
\end{figure*}

\begin{figure*}[t]
  \centering
  \includegraphics[width=1.85\columnwidth]{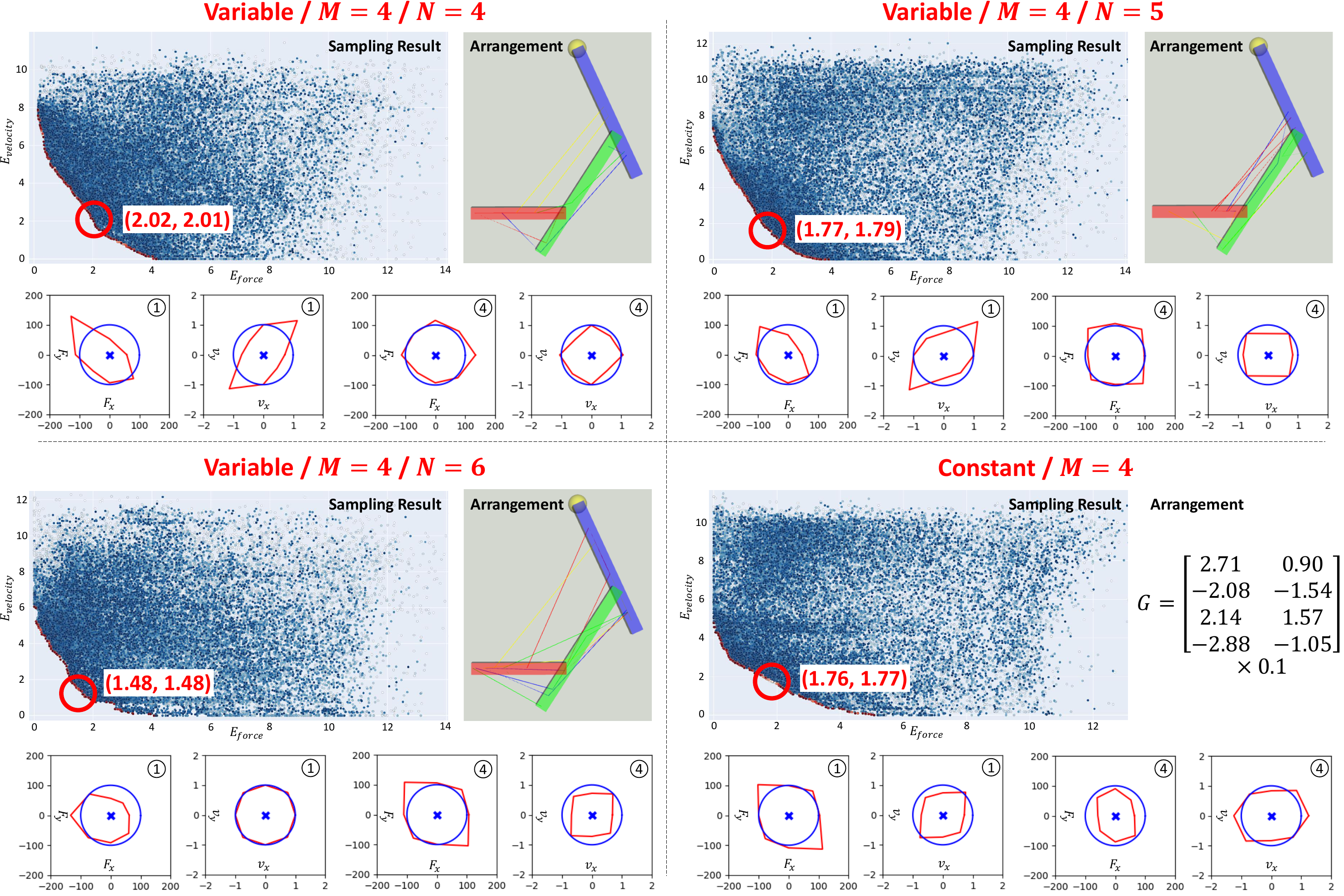}
  \vspace{-1.0ex}
  \caption{The detailed analysis of \textbf{Constant} without restrictions for Target-1 w/o Gravity. For \textbf{Variable} with $M=4$ and $N=\{4, 5, 6\}$ and for \textbf{Constant} with $M=4$, the sampling results and Pareto solutions are shown.}
  \label{figure:exp1c-pareto}
  \vspace{-3.0ex}
\end{figure*}

\subsection{Target-1 w/o Gravity}
\switchlanguage%
{%
  We show the experimental results for \textbf{Target-1} and \textbf{w/o-Gravity}.
  For \textbf{Variable}, the parameters are varied as $M=\{3, 4\}$ and $N=\{2, 3\}$.
  For \textbf{Constant}, the parameters are varied as $M=\{3, 4\}$.
  The results are shown in \figref{figure:exp1-pareto}.

  First, we explain how to interpret the results by referring to the example of \textbf{Variable} when $M=3$ and $N=2$.
  The upper left figure shows the sampling result of optimization, where $E_{force}$ is for the $x$-axis and $E_{velocity}$ is for the $y$-axis.
  The further down to the left of the sampling result graph, the better the solution becomes.
  Among the sampling results, the samples represented by the red dots are the Pareto solutions.
  The solution with the smallest $|E_{force}-E_{velocity}|$ is indicated by a red circle, and $(E_{force}, E_{velocity})$ and its wire arrangement are shown.
  As $M=3$, there are three wires, and as $N=2$, there are two relay points, a start point and an end point.
  Below the figures, the target and feasible OFS and OVS of this design for \ctext{1}--\ctext{4} in \figref{figure:exp-setup} are shown with $F_{\{x, y\}}$ or $v_{\{x, y\}}$ as $\{x, y\}$-axes.
  If the feasible space (red line) exceeds the target space (blue line), $E_{\{force, velocity\}}$ becomes smaller.

  Second, we discuss the overall characteristics of the experiment.
  Since it is possible to increase the operational velocity by keeping the moment arm as small as possible, there is almost certain to be a sampling at $E_{velocity}=0$.
  On the other hand, there are many cases where there is no sampling with $E_{force}=0$, because the moment arm has a maximum value.
  For example, for \textbf{Variable} with $N=2$, or for \textbf{Constant}, there is no sampling at $E_{force}=0$.
  In particular, $E_{force}$ is much larger for \textbf{Constant} than for \textbf{Variable} because it is difficult for \textbf{Constant} to obtain large moment arm (this will be analyzed in detail in the next experiment).
  OFS and OVS show that the velocity takes large values, while the force takes relatively small values.
  Next, $E_{\{force, velocity\}}$ tends to be smaller as the number of wires and relay points increase.
  The larger the number of wires and relay points, the wider the design range becomes, and the more flexibly the target OFS and OVS can be covered.
  For this experimental setting, it is found that increasing the number of relay points is more effective than increasing the number of wires.
  When $N=2$, the feasible OFS and OVS cannot be brought close to the target space, and the feasible space is sharp in a certain direction.
  On the other hand, when $N=3$, it is found that the feasible OFS and OVS are round and close to the target space.
  As for the wire arrangement, when $N=2$ for both $M=3$ and $M=4$, there exist wires that control the first and second joints independently, as well as a wire that controls the two joints simultaneously (to control the second joint independently, the start point of that wire is placed near the first joint at $\textrm {LINK}_{0}$).
  On the other hand, when $N=3$, there are multiple wires that control two joints simultaneously, and we can see diverse wire arrangements.

  Third, a detailed analysis is given for \textbf{Constant}.
  We relax the restrictions of $R^{\{s, e\}}_{d}$ in \figref{figure:exp1-pareto} and show the results when [$R^{s}_{d}$, $R^{e}_{d}$]=[-0.4, 0.4], which is not feasible in practice.
  This allows us to compare \textbf{Variable} and \textbf{Constant} without design restrictions.
  Here, we limit the evaluated joint angles to \ctext{1} and \ctext{4} (the reason will be described later), and \textbf{Variable} with $N=\{4, 5, 6\}$ and $M=4$ is considered for comparison.
  Note that we set $N_{sample}=50000$ for this experiment because the number of parameters is much larger than in \figref{figure:exp1-pareto}.
  The results are shown in \figref{figure:exp1c-pareto}.
  It can be seen that for \textbf{Variable}, better solutions are generated as $N$ is increased.
  For the solution with the smallest $|E_{force}-E_{velocity}|$, the performance of \textbf{Variable} with $N=5$ is almost equal to that of \textbf{Constant}.
  In other words, for \textbf{Constant}, performance without restrictions is much higher than that with restrictions.
  Moreover, by increasing $N$ for \textbf{Variable}, we can obtain higher performance than that of \textbf{Constant}.
  This is because the same $\bm{G}$ is used for each joint angle for \textbf{Constant}, while different $\bm{G}$ is used for each joint angle for \textbf{Variable}.
  On the other hand, the performance of \textbf{Variable} is not as good as or better than that of \textbf{Constant} unless the number of relay points is considerably increased, as in $N=\{5, 6\}$.
  In this experiment, the evaluated joint angles are limited to \ctext{1} and \ctext{4}, which differ greatly in $\bm{J}$.
  If the evaluated joint angles are increased to \ctext{1}--\ctext{4}, it is difficult to form $\bm{G}$ that is appropriate for all of \ctext{1}--\ctext{4}, and even if $\bm{G}$ is always constant or even if $\bm{G}$ is varied, the performance does not change significantly.
  Therefore, even when $N$ is increased, there is no significant difference between \textbf{Constant} and \textbf{Variable}.
}%
{%
  \textbf{Target-1}かつ\textbf{w/o-Gravity}における実験結果を示す.
  ここで, \textbf{Variable}について, $M=\{3, 4\}$, $N=\{2, 3\}$とパラメータを変化させて実験を行っている.
  また, \textbf{Constant}について, $M=\{3, 4\}$とパラメータを変化させて実験を行っている.
  実験結果を\figref{figure:exp1-pareto}に示す.

  まず, \textbf{Variable}, $M=3$, $N=2$の例を参考に図の見方を説明する.
  左上の図は最適化におけるサンプリング結果を示しており, 横軸を$E_{force}$, 縦軸を$E_{velocity}$の値としており, サンプリング結果が左下に行くほど良い解である.
  サンプリング結果のうち, 赤い点で表現されたサンプリングがパレート解である.
  このうち, $|E_{force}-E_{velocity}|$が最も小さい解を赤い丸で示しており, その時の$(E_{force}, E_{velocity})$と設計解を示している.
  この実験では$M=3$のため, 3つのワイヤが存在し, $N=2$のため, 経由点は始点と終止点の2つのみである.
  その下に, この設計における目標値と発揮可能手先力・手先速度空間が図示されている.
  \figref{figure:exp-setup}に示した\ctext{1}--\ctext{4}について, それぞれ$F_{\{x, y\}}$, $v_{\{x, y\}}$を座標系とした図示を行っている.
  このとき, 実際値(赤線)が目標値(青線)を超えていれば, $E_{\{force, velocity\}}$の値が小さくなる.

  次に, その全体的な特徴について述べる.
  まず, モーメントアームをできるだけ小さく取れば手先速度を大きくすることが可能であるため, $E_{velocity}=0$であるサンプリングがほぼ確実に存在する.
  一方で, 設計上モーメントアームには最大値が存在するため, $E_{force}=0$であるサンプリングは存在しない場合も多い.
  例えば, \textbf{Variable}かつ$N=2$のとき, または\textbf{Constant}のときには, $E_{force}=0$であるサンプリングが存在しない.
  特に\textbf{Constant}の場合は, \textbf{Variable}よりもモーメントアームを大きく取りにくいため, $E_{force}$が非常に大きな値となっている(これについては次の実験で詳細な解析を行う).
  手先力・手先速度空間を見ても, 速度は大きな値を取っている一方で, 力の値は比較的小さいことがわかる.
  次に, $(E_{force}, E_{velocity})$の値は, ワイヤ本数や経由点数が大きいほど小さい傾向にある.
  ワイヤ本数や経由点数が増えるほど, 設計の幅が増え, より柔軟に目標手先力・手先速度空間をカバー可能である.
  本実験の設定については, ワイヤ本数を増やすよりも, 経由点数を増やす方が, $E_{force}=0$であるサンプリングが生まれ, その効果が大きいことがわかる.
  手先力・手先速度空間を見ても, $N=2$の場合は発揮可能手先力・手先速度空間を目標値に近づけることが出来ず, その空間が一方方向に尖っている.
  一方で, $N=3$の場合は, 目標値に近く丸い形の発揮可能手先力・手先速度空間を実現できていることがわかる.
  実際の設計については, $N=2$のときは, $M=3$と$M=4$の両者について, 1つ目の関節・2つ目の関節を独立に制御するワイヤと, 2つの関節を同時に制御するワイヤが同じように存在していた(2つ目の関節を独立に制御するために, そのワイヤの始点は$\textrm{LINK}_{0}$における1つ目の関節付近に配置されている).
  一方で$N=3$の場合は, 2つの関節を同時に制御するワイヤが複数存在し, より多様なワイヤ配置を見ることができる.

  最後に, \textbf{Constant}の場合について詳細な解析を行う.
  \figref{figure:exp1-pareto}における$R^{\{s, e\}}_{d}$の制限を緩め, 実際には実現可能ではないが[$R^{s}_{d}$, $R^{e}_{d}$]=[-0.4, 0.4]とした際の結果を示す.
  これにより, 設計的制限が無い場合における\textbf{Variable}と\textbf{Constant}の比較が可能である.
  ここでは評価する関節角度を\ctext{1}と\ctext{4}に限定し(その理由は後に述べる), $M=4$の場合について, \textbf{Variable}における$N=\{4, 5, 6\}$を比較対象とする.
  なお, \figref{figure:exp1-pareto}に比べてパラメータ数が非常に多いため, この実験のみ$N_{sample}=50000$とした.
  その結果を\figref{figure:exp1c-pareto}に示す.
  \textbf{Variable}についは, $N$が増えるほど, より良い解が生成されていることがわかる.
  $|E_{force}-E_{velocity}|$が最小の解については, \textbf{Variable}かつ$N=5$の解と\textbf{Constant}の解のパフォーマンスがほぼ同等であった.
  つまり, 制限を緩めた\textbf{Constant}の性能は, \figref{figure:exp1-pareto}における制限のかかった\textbf{Constant}の性能に比べて非常に高い.
  また, \textbf{Variable}において$N$を増やすことで, \textbf{Constant}よりも高い性能を得ることができる.
  これは, \textbf{Constant}の場合は各関節角度について同じ$\bm{G}$を用いるのに対して, \textbf{Variable}の場合は各関節角度について異なる$\bm{G}$が用いられるためである.
  一方で, $N=\{5, 6\}$と, 経由点数をかなり増やさないと\textbf{Constant}と同程度かそれ以上の性能は得ることは出来ない.
  また, 本実験では評価する関節角度を$\bm{J}$の大きく異なる\ctext{1}と\ctext{4}に制限したが, これを\ctext{1}--\ctext{4}に増やすと, それら全ての関節角度に適切な形の$\bm{G}$を形成することは難しく, $\bm{G}$が常に一定でも, 関節角度ごとに変化しても大きな性能変化は出ない.
  そのため, $N$を大きくしても\textbf{Constant}と\textbf{Variable}の間で大きな違いは出なかった.
}%

\begin{figure*}[t]
  \centering
  \includegraphics[width=1.8\columnwidth]{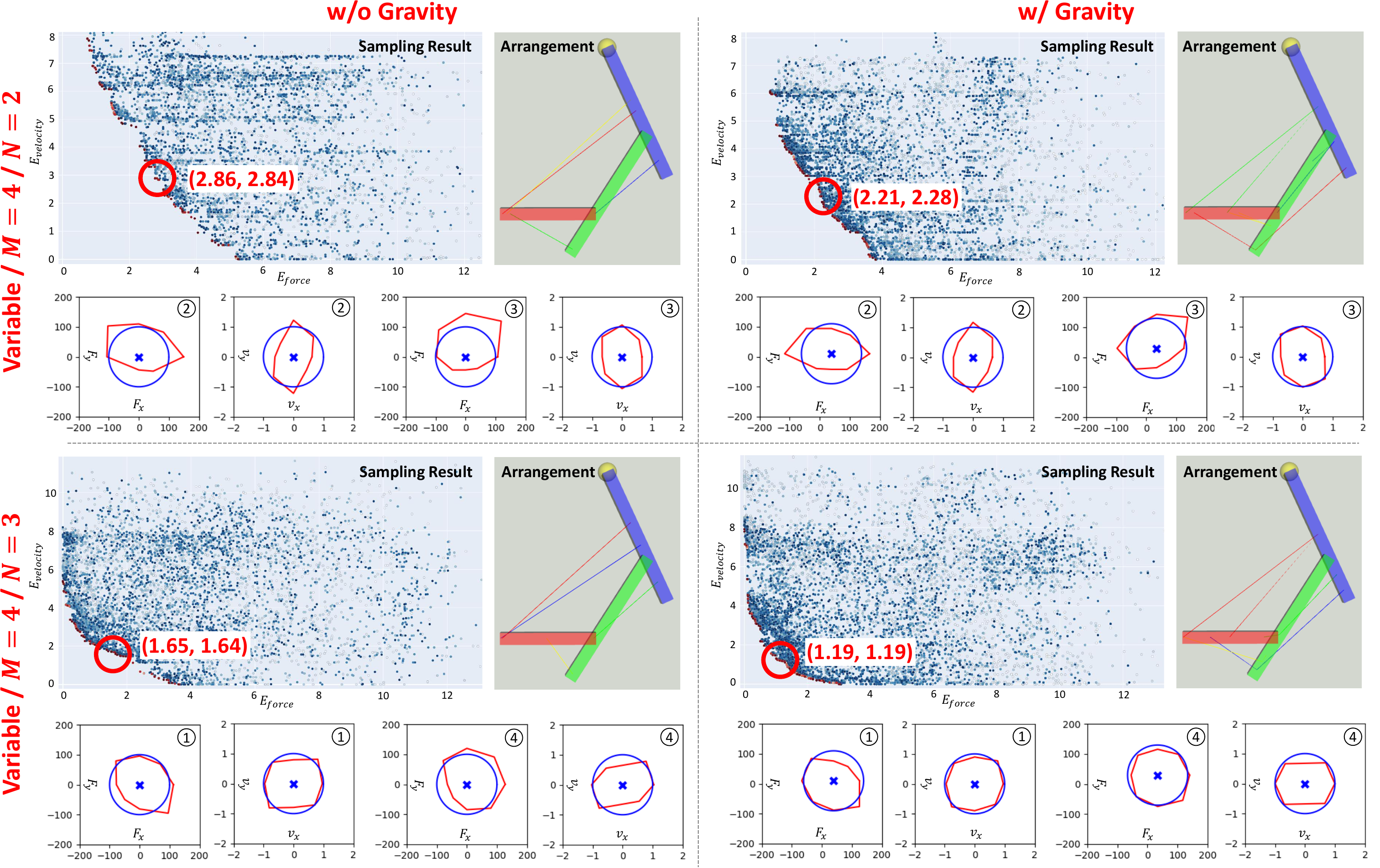}
  \vspace{-1.0ex}
  \caption{The experiment of Target-1 w/ Gravity. For \textbf{Variable} with $M=4$, $N=\{2, 3\}$, and \{\textbf{w/o-Gravity}, \textbf{w/-Gravity}\}, the sampling results and Pareto solutions are shown.}
  \label{figure:exp1g-pareto}
  \vspace{-3.0ex}
\end{figure*}

\subsection{Target-1 w/ Gravity}
\switchlanguage%
{%
  We show the experimental results for \textbf{Target-1} and \textbf{w/-Gravity}.
  We compare the results between \textbf{w/o-Gravity} and \textbf{w/-Gravity} for \textbf{Variable} with $M=4$ by changing the parameters as $N=\{2, 3\}$.
  We limit the evaluated joint angles to \ctext{2} and \ctext{3} for this experiment (the reason will be described later).
  The results are shown in \figref{figure:exp1g-pareto}.
  The performance of the solutions increases by increasing $N$, or with the presence of gravity.
  In particular, the sampling results show that the presence of gravity does not significantly change the minimum value of $E_{force}$, but $E_{velocity}$ becomes smaller overall.
  Gravity always exerts a force in the negative direction on the $y$-axis.
  In the current setup of feasible wire arrangement, the force in the positive direction on the $y$-axis is easily exerted, but the force in the negative direction is hardly exerted and can be compensated by gravity.
  In this experiment, the evaluated joint angles are limited to \ctext{2} and \ctext{3}, which have similar $\bm{J}$.
  If the evaluated joint angles are increased to \ctext{1}--\ctext{4}, the difficulty in forming appropriate $\bm{G}$ for all the joint angles increases, and the performance difference between \textbf{w/o-Gravity} and \textbf{w-Gravity} becomes smaller.
}%
{%
  \textbf{Target-1}かつ\textbf{w/-Gravity}における実験結果を示す.
  本実験では\textbf{Variable}かつ$M=4$のみについて, $N=\{2, 3\}$とパラメータを変化させて\textbf{w/o-Gravity}と\textbf{w/-Gravity}の間で比較実験を行っている.
  また, 評価する関節角度を\ctext{2}と\ctext{3}に限定し実験を行う(その理由は後に述べる).
  実験結果を\figref{figure:exp1g-pareto}に示す.
  $N$を増やすことで, また, 重力が加わることで得られる解の性能がより高くなることがわかる.
  特にサンプリング結果を見ると, 重力が加わることで, パレート解について, $E_{force}$の最小値に大きな変化はないが, $E_{velocity}$が全体的に小さくなっていることがわかる.
  重力によって, $y$軸のマイナス方向への力が常に働く.
  本設定では, 基本的に$y$のプラス方向へ力が出しやすく, マイナス方向への力が出しにくいリンク構造なため, このマイナス方向の力を重力によって補うことができる.
  なお, 本実験では評価する関節角度を$\bm{J}$の似ている\ctext{2}と\ctext{3}に制限したが, これを\ctext{1}--\ctext{4}に増やすと, それら全ての関節角度に適切な形の$\bm{G}$を形成することが難しいことによる影響が大きくなり, \textbf{w/o-Gravity}と\textbf{w-Gravity}の間で大きな性能差が出ない.
}%

\begin{figure*}[t]
  \centering
  \includegraphics[width=1.8\columnwidth]{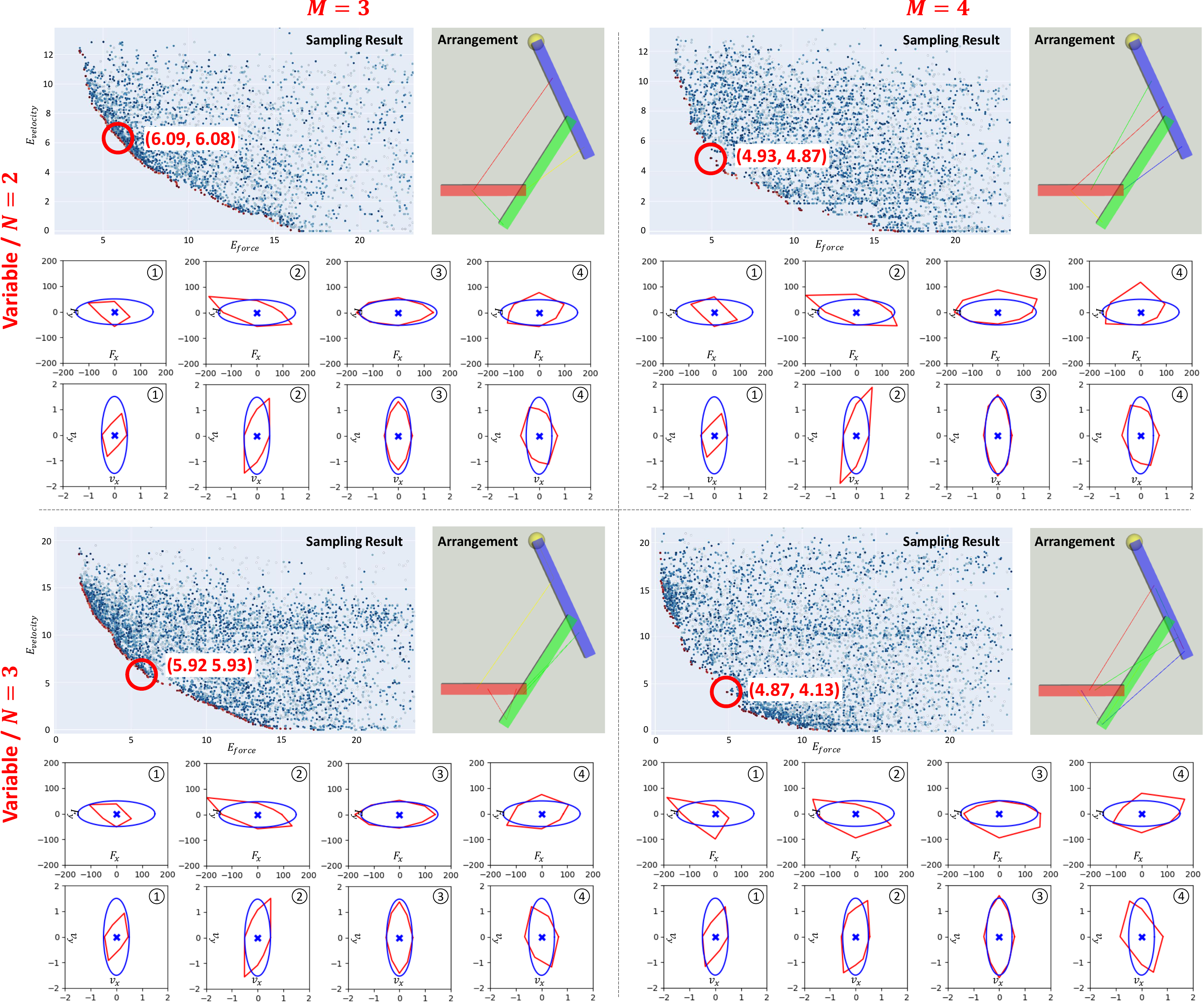}
  \vspace{-1.0ex}
  \caption{The experiment of Target-2 w/o Gravity. For \textbf{Variable} with $M=\{3, 4\}$ and $N=\{2, 3\}$, the sampling results and Pareto solutions are shown.}
  \label{figure:exp2-pareto}
  \vspace{-3.0ex}
\end{figure*}

\subsection{Target-2 w/o Gravity}
\switchlanguage%
{%
  We show the experimental results for \textbf{Target-2} and \textbf{w/o-Gravity}.
  We perform comparative experiments for \textbf{Variable} by changing the parameters as $M=\{3, 4\}$ and $N=\{2, 3\}$.
  The results are shown in \figref{figure:exp2-pareto}.
  It can be seen that the performance of the obtained solutions becomes higher by increasing $M$ or $N$.
  However, increasing $N$ does not significantly change the performance as in \figref{figure:exp1-pareto}.
  Compared to \textbf{Target-1}, for \textbf{Target-2}, the increase in $M$ causes a larger change in performance than an increase in $N$.
  From the result for $N=3$ and $M=4$, the feasible OFS and OVS can realize the sharp target spaces well.
}%
{%
  \textbf{Target-2}かつ\textbf{w/o-Gravity}における実験結果を示す.
  本実験では\textbf{Variable}のみについて, $M=\{3, 4\}$, $N=\{2, 3\}$とパラメータを変化させて比較実験を行っている.
  実験結果を\figref{figure:exp2-pareto}に示す.
  $M$を増やすことで, また, $N$を増やすことで得られる解の性能がより高くなることがわかる.
  一方で, \figref{figure:exp1-pareto}と比較すると, $N$を増やしても得られる解の性能にそれほど大きな変化はない.
  \textbf{Target-1}に比べ, \textbf{Target-2}では$M$の増加の方が$N$の増加よりも大きな性能変化を起こしている.
  $N=3$, $M=4$の結果を見ればわかる通り, 細長い形の目標手先力・手先速度空間にもしっかりその性能が追従していることがわかる.
}%

\section{Discussion} \label{sec:discussion}
\switchlanguage%
{%
  The experimental results are summarized and discussed.
  First, the multi-objective optimization of this study has allowed us to obtain a variety of Pareto solutions where a trade-off exists to achieve the target OFS or OVS.
  It is found that the performance of the solution is improved by increasing the number of wires or the number of relay points for \textbf{Variable}.
  Whether it is better to increase the number of wires or the number of relay points depends on the geometry of the target OFS and OVS.
  Additionally, the performance of \textbf{Constant} is inferior to that of \textbf{Variable}, because \textbf{Constant} cannot have large moment arm.
  On the other hand, by removing the restrictions of the moment arm, higher performance can be obtained.
  Since \textbf{Constant} uses the same $\bm{G}$ for all joint angles, \textbf{Variable} theoretically has better performance as it can change $\bm{G}$ according to the joint angle.
  However, in order for \textbf{Variable} to achieve higher performance than that of \textbf{Constant} without restrictions, it is necessary to significantly increase the number of relay points.
  This implies that a large degree of freedom in wire arrangement is necessary to make $\bm{G}$ nonlinear enough to accommodate changes in $\bm{J}$ at each joint angle.
  Note that by increasing the gear ratio of the motor, \textbf{Constant} can achieve high performance without relaxing the restrictions, but at the cost of backdrivability.
  For \textbf{Variable}, there is also a trade-off that the friction at the pulleys should increase as the number of relay points increases.
  In addition, the effect of gravity may have a positive effect depending on the setting.
  The wire drive can produce anisotropic torque depending on the direction of joint rotation, and so gravity can be used effectively in many cases.

  We discuss future issues.
  First, the design parameters of the manipulator in this study were relatively simple because we handled a planar manipulator.
  When the manipulator becomes 3D, it is necessary to take into account the interference of wires during movement, which requires a more complicated formulation.
  In addition, the joint structure is not optimized in this study.
  Although simultaneous optimization of link lengths with wire arrangements can be performed in the same framework, the design parameter space becomes huge when closed links and branching of links are considered.
  Furthermore, it should be noted that aspects such as wire elongation and friction are not addressed in this study.
  Second, although we have considered the feasible OFS and OVS as the objective function in this study, this framework can flexibly consider a wider variety of objective functions.
  In the future, it is necessary to consider more complex and realistic settings for objective functions in accordance with the actual tasks to be realized.
  In addition, we would like to develop a tendon-driven robot that actually implements the obtained solution, and confirm its effectiveness.
}%
{%
  得られた実験結果についてまとめ, 考察する.
  まず, 本研究の多目的最適化によって, 目標手先力・手先速度空間の実現というトレードオフのある多様なパレート解を得ることができた.
  \textbf{Variable}について, ワイヤ本数や経由点数を増やすことで, 手先力・手先速度空間における解の性能が向上することがわかった.
  この際, ワイヤ本数と経由点のどちらを増やしたほうが良いかについては, 目標手先力・手先速度空間の形状に依存して異なる.
  \textbf{Constant}については, \textbf{Variable}に比べモーメントアームを大きく取ることができないため, その性能は劣る.
  一方で, モーメントアームの制約を除くことで, 高い性能を得ることができる.
  \textbf{Constant}はどの関節角度についても同じ$\bm{G}$を用いるため, 理論的には関節角度に応じて$\bm{G}$を変更可能な\textbf{Variable}の方が性能が高い.
  しかし, \textbf{Variable}が制約を除いた\textbf{Constant}よりも高い性能を出すためには, 経由点の数を大きく増やす必要がある.
  これは, 各関節角度における$\bm{J}$の変化に対応できるほど非線形な$\bm{G}$を得るには大きなワイヤ配置自由度が必要であることを示唆している.
  なお, モータのギア比を高くすることで, \textbf{Constant}は制約を緩和しなくても高い性能を出すことができるが, その分バックドライバビリティが犠牲となる.
  もちろん, \textbf{Variable}についても経由点を増やすほどプーリにおける摩擦が増加するはずであり, トレードオフが存在する.
  また, 設定によっては重力の効果がプラスに働く場合がある.
  特にワイヤ駆動は関節の回転方向に応じてトルクに異方性を出すことが可能であり, 重力の効果が利用できる場合が多い.

  今後の課題について述べる.
  まず設計パラメータについて, 本研究では平面マニピュレータを扱ったため, 比較的シンプルに設計パラメータを構築することができた.
  一方で, マニピュレータが3次元になったとき, 動作時におけるワイヤの干渉を考慮に入れる必要があるため, より複雑な定式化が必要である.
  また, 本来はワイヤの伸びや摩擦なども扱うべきであるが, 本研究では扱えていない.
  また, 本研究では関節構造の最適化は行っていない.
  リンク長の同時最適化程度であれば同じフレームワークで可能だが, 関節構造に閉リンクや枝分かれを考えたとき, その設計パラメータ空間は膨大なものとなり, パラメタライズが難しくなる.
  今後より複雑な身体構造の最適化についても扱いたい.
  次に評価関数について, 本研究では発揮可能手先力・手先速度空間について考えたが, 本フレームワークでは, より多様な評価関数を柔軟に与えることができる.
  今後実際に実現したいタスクに合わせ, より複雑で現実的な設定のもと, 評価値を考えていく必要がある.
  また, 得られた解を実際に実装したワイヤ駆動ロボットを開発し, その有効性を確認していきたい.
}%

\section{CONCLUSION} \label{sec:conclusion}
\switchlanguage%
{%
  We proposed a design optimization method for wire arrangement in a tendon-driven robot with variable relay points.
  While previous studies have focused on conditions with constant moment arm or fixed links to attach relay points to, this study deals with a broader problem and performs multi-objective black-box optimization aiming at achieving a target operational force space and velocity space.
  Various designs can be represented by setting the links each wire attaches to and the positions of the relay points as variables.
  A body using variable relay points can be designed more freely in terms of its physical capabilities compared to a general tendon-driven robot with constant moment arm using pulleys.
  We expect that this concept will be a key to solving various wire arrangement design problems.
}%
{%
  本研究では, 可変ワイヤ経由点を持つワイヤ駆動ロボットにおけるワイヤ配置の設計最適化手法を提案した.
  これまではモーメントアーム一定や経由点を取り付けるリンクが固定な研究がなされてきたが, 本研究ではより広い問題を扱い, 目標手先力・手先速度空間の実現を目指した多目的ブラックボックス最適化を行った.
  各ワイヤにおける経由点の取り付けリンク・その位置を変数とすることで, 多様な設計を表現することができる.
  本研究のように可変ワイヤ経由点を持つモーメントアーム可変な身体は, プーリを用いたモーメントアーム一定の一般的なワイヤ駆動に比べ, その身体能力をより自由に設計することが可能である.
  本手法の考え方が今後の多様なワイヤ配置設計問題を解く鍵となることを期待する.
}%

{
  \bibliographystyle{IEEEtran}
  \bibliography{main}
}

\end{document}